\pdfoutput=1
\documentclass{article}
\usepackage[nusbwlab]{iclr2025_conference}
\usepackage{times}
\bwlheadplain
\renewcommand{\bwlheadkind}{Preprint}

\usepackage{amsmath,amssymb,amsfonts}

\usepackage{amsmath,amsfonts,bm}

\def\eqref#1{equation~\ref{#1}}

\def\1{\bm{1}}

\DeclareMathAlphabet{\mathsfit}{\encodingdefault}{\sfdefault}{m}{sl}
\SetMathAlphabet{\mathsfit}{bold}{\encodingdefault}{\sfdefault}{bx}{n}

\usepackage{url,booktabs,nicefrac,microtype,xcolor,colortbl,enumitem}
\usepackage{graphicx,multirow,algorithm,algorithmic,float,tcolorbox,listings}
\usepackage{placeins,needspace,hyperref}
\hypersetup{colorlinks=true,linkcolor=bwlblue,citecolor=bwlblue,urlcolor=bwlblue,
  pdftitle={MolLean: A Process-Verifiable Language for Chemical Reasoning},
  pdfauthor={Jiatong Li and Yatao Bian}}


\newcommand{\op}[1]{\texttt{#1}}

\definecolor{backred}{RGB}{255,190,190}
\definecolor{backblue}{RGB}{220,230,250}
\definecolor{backgreen}{RGB}{224,244,224}
\graphicspath{{figures/}}
\lstdefinestyle{trace}{basicstyle=\footnotesize\ttfamily,breaklines=true,
  columns=fullflexible,keepspaces=true,frame=none,xleftmargin=0.5em}

\title{MolLean: A Process-Verifiable Language for Chemical Reasoning}
\author{Jiatong Li \& Yatao Bian\textsuperscript{\dag} \\
National University of Singapore \\
\textsuperscript{\dag}Corresponding to: \texttt{ybian@nus.edu.sg} \\
\url{https://github.com/phenixace/Open-MolLean}}

\begin{document}
\maketitle
\begin{abstract}
Chemical chain-of-thought can contain hallucinated structural and property claims. We thereby introduce MolLean, a process-verifiable language that grounds chemical reasoning in an explicit molecular state. 
A layered Lean-based kernel checks state transitions and assertions over molecular readouts, from primitive graph edits to compound transformations. Meanwhile, a typed witness protocol separates native structural checks from externally supplied measurements. 
We further introduce MolLean-MCP, a stateful tool interface that exposes these transitions, checks, and readouts during generation: the model receives feedback to decide its next action, and rejected actions leave the state unchanged. MolLean supports one-pass trace execution, whole-trace refinement, and interactive reasoning through MolLean-MCP. We evaluate these three modes with direct chain-of-thought on 8,800 matched examples across 12 tasks spanning molecular generation, editing, optimization, and reactions. With Qwen3.6-27B, whole-trace refinement outperforms direct chain-of-thought on all 12 tasks. MolLean-MCP complements whole-trace refinement with stepwise verification feedback, reaching better performance in reaction tasks. Together, these results highlight an executable foundation for chemical reasoning, in which explicit chemical commitments can be checked and revised rather than remaining solely in free-form explanations.
\end{abstract}

\section{Introduction}
\label{sec:introduction}
Chain-of-thought prompting makes a model's intermediate reasoning visible \citep{wei2022chain}. In chemistry, this is appealing as chemists could examine the intermediate reasoning steps to improve the explainability of the final answer \citep{wang2026chem,narayanan2026training,li2025mol,zhao2025molreasoner}. However, a model can invoke a functional group absent from a molecule, lose track of a bond, or claim a property improvement without establishing it. Analyses of chemical reasoning find fabricated structural claims even alongside correct answers \citep{li2026scratchpad}, and broader evaluations reveal uneven chemical competence and overconfidence \citep{mirza2025chembench}. These errors concern the stated chemistry itself, a failure mode complementary to causal unfaithfulness, where an explanation misrepresents how the answer was produced \citep{turpin2023unfaithful}.

We therefore ask \emph{what must a chemical reasoning language represent for its steps to be verifiable?} Lean offers a precedent: a model proposes a derivation and a proof assistant checks its formal obligations \citep{demoura2021lean4,jiang2023draft,yang2023leandojo}. Chemical reasoning combines graph transformations, structural observations, empirical estimates, and choices among plausible candidates. Chemistry tools already support capable agents \citep{bran2024chemcrow,boiko2023coscientist}; the key step is to give successive operations and claims a shared molecular referent and an explicit acceptance condition.

We introduce \textbf{MolLean}, a process-verifiable language grounded in a maintained molecular state (Figure~\ref{fig:overview}). A \emph{transition} constructs, modifies, or reorganizes represented molecules. A \emph{readout} obtains information without changing them, and an assertion tests a claim against that information. The model selects actions and obligations; a Lean-based kernel checks admissible transitions and native structural predicates. Descriptor- and knowledge-dependent assertions obtain typed witnesses from external providers. These distinctions let construction, transformation, and observation share one language while each keeps its own form of evidence.

The state does more than record the output: it determines which atoms a later edit addresses, which molecule a property readout measures, and whether an accepted prefix remains usable after a failed proposal. MolLean supports primitive edits, construction macros, joint transformations, and candidate introduction at different granularities. An accepted edit supplies the referent for subsequent checks, and a rejected edit leaves the maintained state intact. Verification thus applies directly to the model's declared commitments.

We expose the same language through whole-trace and interactive interfaces. \textbf{MolLean One-pass} submits a complete trace; \textbf{MolLean Refinement} revises a rejected trace from the initial state. \textbf{MolLean-MCP} lets the model inspect the current state, attempt an operation, and use immediate feedback to choose its next action. External feedback is central to revision: tool-grounded criticism guides reasoning more reliably than intrinsic self-correction \citep{huang2024selfcorrect,gou2024critic}. MolLean specifies what that feedback refers to and when it commits a change.

Our contributions are as follows:
\begin{itemize}[leftmargin=*,topsep=0pt,itemsep=1pt,parsep=0pt]
\item \emph{A state-grounded chemical reasoning language and verification contract.} MolLean unifies molecular transitions, readouts, and assertions, with explicit boundaries between native structural checks and externally witnessed claims.
\item \emph{Two feedback regimes over shared execution semantics.} Whole-trace refinement revises complete traces from the input; MolLean-MCP interacts with the maintained state and commits only accepted edits.
\item \emph{An evaluation on 8,800 matched examples across 12 tasks.} With Qwen3.6-27B, refinement improves over direct chain-of-thought on every task, and MCP reaches the highest reaction accuracy. On ChEBI-20, refined traces reach 79.0\% soundness and 33.1\% answer accuracy, identifying task fidelity as the next challenge beyond repairing rejected steps. Qwen3-4B gains substantially on structural tasks.
\end{itemize}

\section{MolLean}
\label{sec:model}
\begin{figure}[t]
\centering
\includegraphics[width=\linewidth]{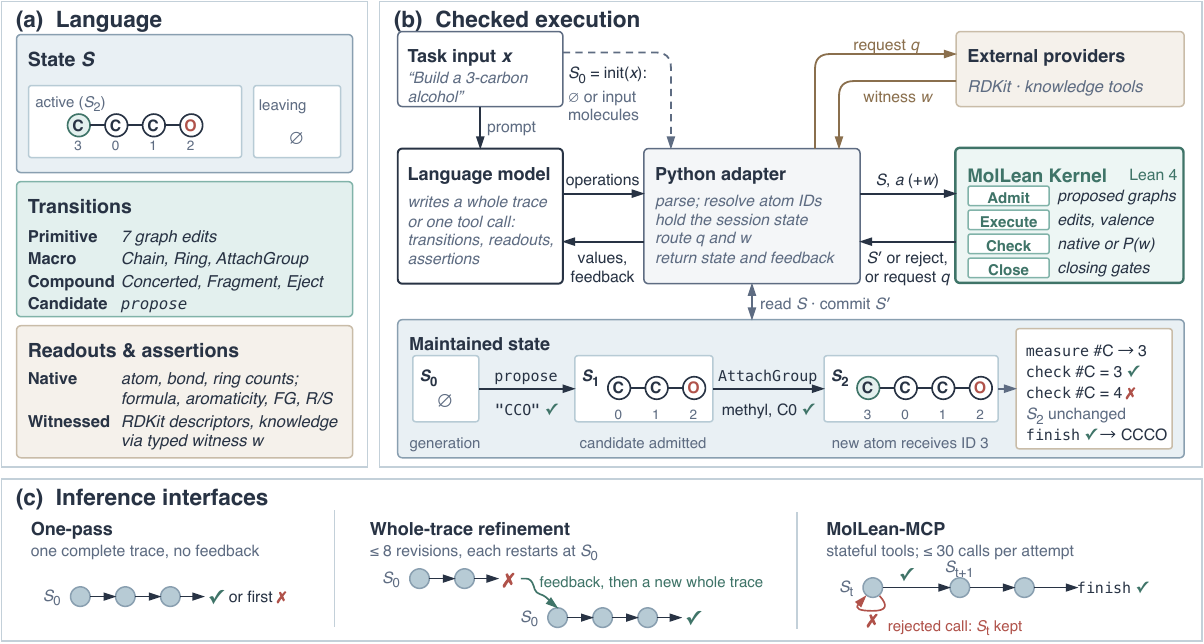}
\caption{\textbf{MolLean's language, execution contract, and inference interfaces.} (a) The state, transition families, and read-only queries; native predicates are decided in Lean, while witnessed claims apply a Lean predicate to a typed provider response. (b) The adapter holds $S_0=\operatorname{init}(x)$, sends each operation to the Lean kernel, and commits only accepted successors; witness requests and responses pass through the adapter. In the example, generation starts empty, admits ethanol as $S_1$, and attaches methyl to obtain 1-propanol as $S_2$; a false carbon count is rejected and $S_2$ is unchanged. (c) One-pass executes one trace; refinement rewrites a rejected trace from $S_0$; MCP continues from the current state. Dataset references are used only by an independent scorer.}
\label{fig:overview}
\end{figure}

MolLean is a reasoning language built around a simple contract: an operation refers to the current molecular state, verification decides whether it is admissible, and only an accepted transition commits a new state. Readouts and assertions observe that same state without editing it. Lean implements the chemical core, while Python handles parsing, state transport, reference resolution, and inference control (Figure~\ref{fig:overview}).

\subsection{A Language Grounded in Molecular State}
Execution begins at $S_0=\operatorname{init}(x)$ for task input $x$. The state contains active molecules, a leaving pool, atom references, and bookkeeping counters. Molecules are graphs labelled with elements, bonds, charges, radicals, and aromaticity. The leaving pool retains fragments that are excluded from the answer and tracked in bookkeeping. Generation can start empty; other tasks load the supplied molecules.

A \emph{transition} attempts to change the state. A \emph{readout} observes it without changing its molecular contents, and an assertion constrains that readout, for example by requiring a count or property improvement. Complete traces express these constraints; the interactive interface also exposes value-returning queries. Edits form one class of transitions, alongside construction, candidate introduction, and fragment separation.

Let $a_t$ be a declared operation, $\Gamma$ the verifier configuration, and $W$ the external provider. We write one execution step as
\begin{equation}
V_{\Gamma,W}(S_{t-1},a_t)=
\begin{cases}
\operatorname{accept}(S_t,o_t) & \text{if } a_t \text{ passes its checks in } S_{t-1},\\
\operatorname{reject}(r_t) & \text{otherwise; the current state remains } S_{t-1},
\end{cases}
\label{eq:transition}
\end{equation}
where $o_t$ is the observation or verdict and $r_t$ is the reason for rejection. Accepted readouts and assertions return $S_t=S_{t-1}$, so only accepted transitions commit a new state. Only declared operations enter this contract.

Joining molecules merges their atom-index spaces. Whole-trace execution remaps later references after tracked primitive merges; MCP supplies the current atom map for subsequent calls.

\subsection{State Transitions at Different Granularities}
Seven primitives add or remove atoms, form or break bonds, and update bond order, charge, or radicals. They check preconditions and affected valences. Intermediates may be under-saturated to permit construction, and connectivity is enforced by the relevant operation or terminal policy.

Construction macros such as \op{Chain}, \op{Ring}, and \op{AttachGroup} reduce index manipulation. Compound actions express joint bond changes, proton transfer, and fragment separation. \op{Fragment} splits a bridge bond into active molecules; \op{Eject} moves a separated component into the leaving pool. Named reactions expand into these operations. Most expansions execute in Lean; Python fragment adapters submit operations for the same checks.

Granularity determines the verification unit. A \op{Concerted} bundle applies edits atomically and checks the resulting valences once all edits apply. \op{propose} introduces a structurally checked candidate in a single step.

Bookkeeping covers graph atoms, formal charge, and radical parity across both pools. Addition, removal, and candidate introduction update the ledger accordingly. Charge and radical balance are closing obligations. Terminal \op{realizability} checks valence, connectivity, aromaticity, and closed-shell rules.

\subsection{State Readouts and External Witnesses}
A readout associates a query with the state being examined. MolLean supports native checks for seven structural families: atom counts, bond counts, ring counts, molecular formula, aromaticity, built-in functional-group patterns, and retained R/S stereocenter annotations. Element counts include implicit as well as explicit hydrogen, while molecular-formula checks require all present elements. Ring counts use a graph-theoretic definition.

MCP distinguishes \op{measure}, which returns a value, from \op{check}, which tests a proposed value. The reader obtains simple quantities from serialized state and resolves pattern-sensitive counts through Lean predicates, returning a value only when a compatible one is confirmed.

RDKit supplies descriptors, fingerprints, and molecular comparisons \citep{rdkit}; knowledge providers supply lookup-dependent information. For an assertion $a$, a typed witness protocol separates obtaining evidence from checking the claim:
\begin{equation}
q=\operatorname{request}(S,a),\qquad w=W(q),\qquad
\operatorname{accept}(a)\iff P_a(w)=1.
\label{eq:witness}
\end{equation}
Lean requests typed fields and applies the predicate. Built-in native predicates compute directly; other supported functional-group patterns use RDKit counts.

For \op{PropertyChange} and \op{FGCountDelta}, the provider binds the baseline to the task's initial molecule. These checks use only the public task specification.

\subsection{Process Soundness and Task Fidelity}
For a parsed MolLean trace $\tau=(a_1,\ldots,a_T)$ and closing obligations $G_\Gamma$, we define \emph{trace soundness} operationally:
\begin{equation}
s_{\Gamma,W}(\tau;x)=\mathbf{1}[
\text{all declared trace operations are accepted and }G_\Gamma(S_T)=1].
\label{eq:soundness}
\end{equation}
Missing or unparseable traces count as unsound. For MCP, soundness applies to the accepted trace, excluding read-only queries and rejected proposals.

This contract concerns the represented process. \emph{Task fidelity} asks whether the initial representation, selected referents, and expressed obligations capture the input instruction. For example, a trace that verifiably lowers QED is sound with respect to its own obligations and unfaithful to an instruction to raise QED: it executes the wrong objective correctly.

Answer correctness $c(\hat y;x,y)$ is a further, independently scored outcome, where $\hat y$ is obtained from the reached state and $y$ denotes any held-out reference. Thus
\begin{equation}
s_{\Gamma,W}(\tau;x)=1\;\not\Rightarrow\;c(\hat y;x,y)=1.
\label{eq:separation}
\end{equation}
We therefore report process soundness and task scores as separate measures.

The implementation combines proofs and executable checks. Lean proves soundness properties for valence and connectivity validation, a conditional aromaticity result, and six native assertion boundaries (Appendix~\ref{app:implementation}).

\subsection{Whole-Trace and Interactive Inference}
\textbf{MolLean One-pass} executes a generated complete trace until acceptance or the first rejection. \textbf{MolLean Refinement} starts from that same completion, returning rejection feedback and indexed state to the model. It revises rejected traces within a fixed budget, restarting each at $S_0$ and stopping once accepted. Thus the model can replace its entire plan after a failure.

\textbf{MolLean-MCP} exposes transitions, checks, and readouts as stateful tools, supplied to the model as function-calling schemas within the inference loop. The model can inspect the state before choosing its next operation; rejected proposals leave it unchanged. \op{finish} invokes closing gates. MCP uses the language interactively; One-pass and Refinement submit complete programs.

Verification compares only against the input or a self-proposed candidate; dataset references enter only the independent scorer (Section~\ref{sec:experiments}). Each task declares its initialization, allowed operations, closing obligations, and scorer.

\section{Experiments}
\label{sec:experiments}
We ask whether MolLean improves task performance, how accepted traces relate to correct answers, and when interactive verification is useful. The main comparison uses Qwen3.6-27B; an additional Qwen3-4B evaluation examines whole-trace inference with a smaller model.

\subsection{Tasks, Protocols, and Metrics}
All protocols are evaluated on the same 8,800 examples across 12 task types: 500 examples for each of nine S$^2$-Bench tasks \citep{li2026speak}, 500 forward and 500 retrosynthesis examples from USPTO-50k \citep{schneider2016role}, and the complete 3,300-example ChEBI-20 test set \citep{edwards2021text2mol,edwards2022molt5}. Reaction classes are supplied to every protocol, and USPTO uses a deterministic reaction-level split. Invalid answers and unsuccessful episodes remain in the denominator.

\paragraph{Protocols.}
\textbf{Direct CoT} generates free-form reasoning and a final SMILES answer. \textbf{MolLean One-pass} emits one complete trace without feedback. \textbf{MolLean Refinement} starts from the same one-pass completion and allows up to eight whole-trace refinements. \textbf{MolLean-MCP} constructs a trace through stateful tools, with up to two attempts and 30 counted calls per attempt. We abbreviate these protocols as CoT, One-pass, Refine, and MCP. All MolLean results use the same verifier version; Appendix~\ref{app:settings} details the inference settings.

\paragraph{Metrics.}
S$^2$-Bench generation and editing use structural-count and functional-group-change success criteria. For optimization, we report success-weighted input similarity,
\begin{equation}
\operatorname{score}_{\rm opt}(\hat y,x)
=\mathbf{1}[\text{requested property improves}]
\operatorname{Tanimoto}(f(\hat y),f(x)),
\label{eq:opt}
\end{equation}
where $f$ is a radius-2, 2,048-bit Morgan fingerprint \citep{rogers2010ecfp}. Direction-only success is reported separately in Appendix~\ref{app:tasks}. ChEBI-20 compares the first InChIKey block, which encodes molecular connectivity \citep{heller2015inchi}; reactions require equality of the complete sets of component skeleton keys. We report percentages for binary scores and $100\times$ Equation~\ref{eq:opt} for optimization, task by task. Trace soundness (Equation~\ref{eq:soundness}) is reported alongside answer correctness.

\begin{table}[t]
\centering
\caption{\textbf{Qwen3.6-27B task performance on matched examples.} Refine permits eight whole-trace revisions; MCP uses interactive verification. Bold marks the highest score. $\dagger$ denotes $100\times$ success-weighted similarity.}
\label{tab:performance}
\small
\setlength{\tabcolsep}{6pt}
\begin{tabular}{lrrrrr}
\toprule
Task & $N$ & CoT & One-pass & Refine & MCP \\
\midrule
Atom counts & 500 & 42.0 & 61.4 & \textbf{88.8} & 82.6 \\
Bond counts & 500 & 20.8 & 39.8 & \textbf{69.0} & 56.2 \\
Functional groups & 500 & 25.6 & 38.8 & \textbf{41.8} & 37.2 \\
Add component & 500 & 55.8 & 93.6 & \textbf{100.0} & \textbf{100.0} \\
Delete component & 500 & 85.0 & 94.6 & \textbf{97.0} & \textbf{97.0} \\
Substitute component & 500 & 81.0 & 94.2 & \textbf{96.0} & 95.4 \\
logP optimization $\dagger$ & 500 & 24.0 & 65.7 & 69.5 & \textbf{70.1} \\
MR optimization $\dagger$ & 500 & 22.5 & 60.2 & \textbf{66.1} & 65.6 \\
QED optimization $\dagger$ & 500 & 24.8 & 41.1 & \textbf{59.0} & 49.4 \\
Forward reaction & 500 & 35.0 & 23.6 & 62.8 & \textbf{66.4} \\
Retrosynthesis & 500 & 3.0 & 11.6 & 20.8 & \textbf{27.4} \\
ChEBI-20 text to molecule & 3,300 & 18.1 & 25.3 & \textbf{33.1} & 23.5 \\
 
\bottomrule
\end{tabular}
\end{table}

\subsection{Whole-Trace Refinement Improves Task Performance}
Refinement improves over Direct CoT on all 12 task types (Table~\ref{tab:performance}), with gains of 12.0--48.2 percentage points under each task's binary success criterion. For optimization, this range uses direction-only success, and similarity-weighted scores show the same gains. Atom and bond counts improve by 46.8 and 48.2 points, and component addition by 44.2. Paired bootstrap intervals for the primary Refine-minus-CoT score exclude zero in every task (Table~\ref{tab:verification}).

One-pass already improves over CoT on 11 tasks. On forward reactions, One-pass (23.6\%) trails CoT (35.0\%), and refinement raises accuracy to 62.8\%, showing that feedback is most valuable when the initial plan misses the verifier's obligations. Improvements extend to reference-based tasks, with ChEBI-20 accuracy increasing from 18.1\% to 33.1\% and retrosynthesis from 3.0\% to 20.8\%. These results identify two complementary benefits: expressing a candidate as an executable trace, and using a failed check to revise that candidate.

\paragraph{Recovery versus loss.}
The shared one-pass completion allows a paired comparison. If $R$ counts unsuccessful one-pass answers that become successful after refinement and $L$ counts successes that become failures, then
\begin{equation}
\operatorname{Acc}_{\rm refine}-\operatorname{Acc}_{\rm one\text{-}pass}=(R-L)/N.
\label{eq:recovery}
\end{equation}
Forward prediction recovers 207 answers and loses 11; retrosynthesis recovers 48 and loses two. ChEBI-20 recovers 290 and loses 33, for a net gain of 257 correct answers. The full paired counts appear in Appendix~\ref{app:paired}. Among initially unsuccessful answers, the recovered fraction is 54.2\% for forward prediction, 10.9\% for retrosynthesis, and 11.8\% for ChEBI-20. The remaining failures concentrate on tasks that require identifying a particular reference structure.

\begin{figure}[t]
\centering
\includegraphics[width=\linewidth]{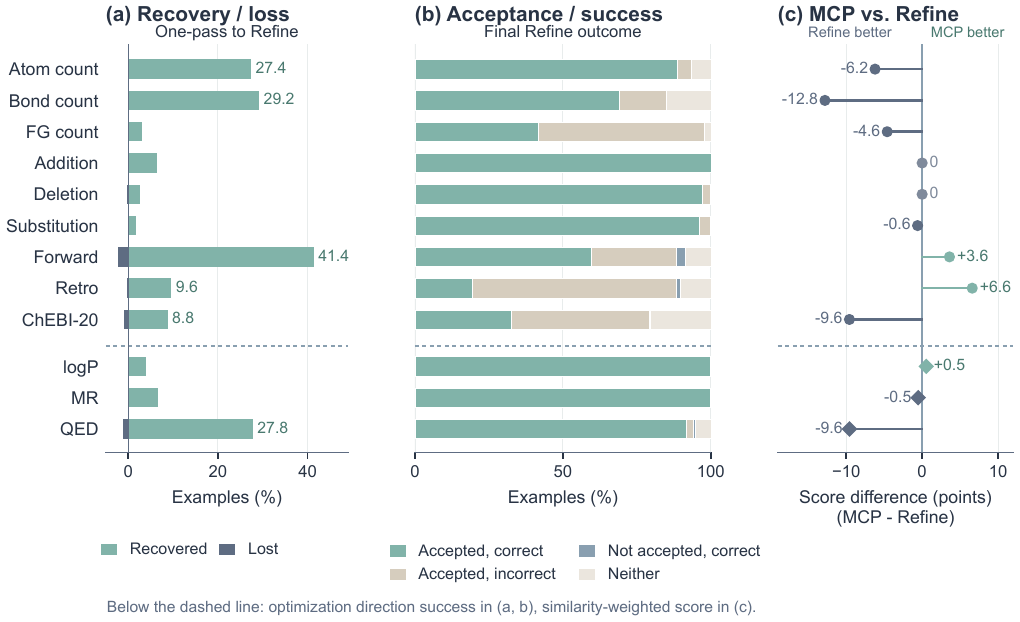}
\caption{\textbf{Repair, specification gaps, and interactive verification.} (a) Refine recovers more One-pass failures than it loses successes, as fractions of each task's examples. (b) Joint Refine outcomes separate accepted traces from correct answers. (c) MCP-minus-Refine primary-score differences show task-dependent benefits. Optimization appears below the dashed line: (a,b) use direction-only success, while (c) uses similarity-weighted score points. Other task differences are percentage points.}
\label{fig:analysis}
\end{figure}

Recovery outweighs loss on every task (Figure~\ref{fig:analysis}a). Accepted one-pass traces keep their answers, as the refinement stopping rule prescribes, so feedback concentrates on repairing rejected commitments.

\begin{figure}[t]
\centering
\includegraphics[width=\linewidth]{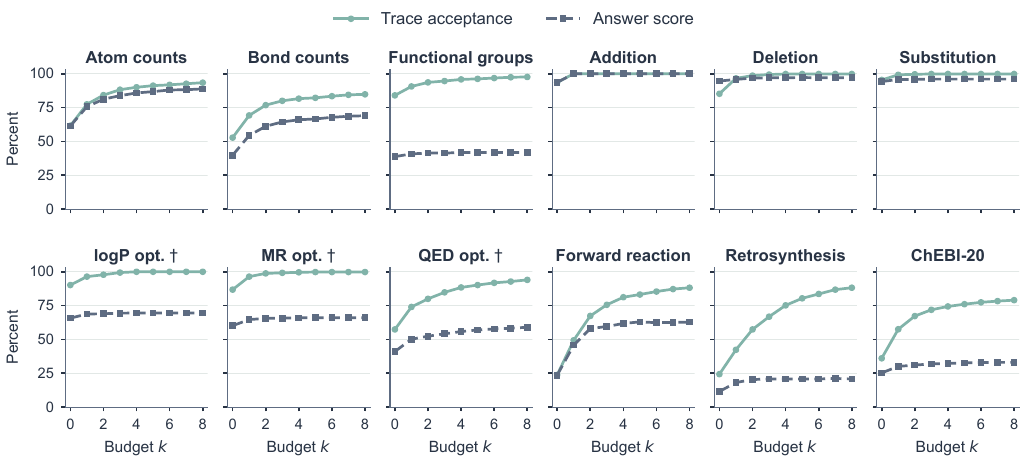}
\caption{\textbf{Whole-trace refinement across revision budgets for all 12 tasks.} Each panel uses the complete Qwen3.6-27B task cohort ($N=500$; ChEBI-20 $N=3{,}300$). Solid curves show trace acceptance and dashed curves the primary answer score; $\dagger$ marks $100\times$ success-weighted similarity. $k=0$ is One-pass and $k=8$ is the Refine column of Table~\ref{tab:performance}.}
\label{fig:refinement-rounds}
\end{figure}

\paragraph{Revision budget.}
Figure~\ref{fig:refinement-rounds} tracks acceptance and answer score after at most $k$ revisions for all 12 tasks, with $k=0$ the shared One-pass completion. Most gains arrive within two revisions: forward-reaction accuracy rises from 23.6\% at $k=0$ to 58.0\% at $k=2$ and 62.8\% at $k=8$, and component addition reaches 100.0\% by $k=2$. On structural generation, acceptance and answer score rise together, with atom counts reaching 93.4\% acceptance and 88.8\% accuracy. On retrosynthesis and ChEBI-20, acceptance keeps rising between $k=2$ and $k=8$, from 57.4\% to 88.2\% and from 67.3\% to 79.0\%, while accuracy moves from 20.4\% to 20.8\% and from 31.1\% to 33.1\%. Later revisions on these tasks mainly repair expressed obligations, which Section~\ref{sec:fidelity} examines.


\subsection{Soundness Separates Execution from Task Fidelity}
\label{sec:fidelity}
Figure~\ref{fig:analysis}b separates acceptance from task success. Table~\ref{tab:verification} reports soundness for every protocol. Refinement raises soundness over One-pass on all 12 tasks, and MCP matches or exceeds Refinement on 10 of them, reaching 98.6\% on forward reactions and 98.0\% on retrosynthesis. The relation between soundness and success depends on what the task asks the model to express.

\begin{table}[t]
\centering
\caption{\textbf{Qwen3.6-27B trace soundness and uncertainty.} Soundness is the percentage of examples with an accepted trace (Equation~\ref{eq:soundness}); intervals are paired 95\% bootstrap intervals (4,000 resamples) for the primary Refine-minus-CoT score.}
\label{tab:verification}
\small
\setlength{\tabcolsep}{6pt}
\begin{tabular}{lrrrr}
\toprule
& \multicolumn{3}{c}{Soundness (\%)} & Refine $-$ CoT \\
Task & One-pass & Refine & MCP & 95\% interval \\
\midrule
Atom counts & 61.6 & 93.4 & 96.6 & [42.2, 51.6] \\
Bond counts & 52.8 & 84.8 & 90.4 & [43.8, 52.8] \\
Functional groups & 84.0 & 97.6 & 98.2 & [12.8, 19.6] \\
Add component & 93.6 & 100.0 & 100.0 & [39.6, 48.6] \\
Delete component & 85.2 & 99.8 & 98.8 & [8.8, 15.2] \\
Substitute component & 95.4 & 99.8 & 100.0 & [12.0, 18.2] \\
logP optimization & 90.2 & 100.0 & 100.0 & [42.7, 48.4] \\
MR optimization & 86.8 & 99.8 & 100.0 & [40.7, 46.4] \\
QED optimization & 57.4 & 94.0 & 92.8 & [31.3, 37.1] \\
Forward reaction & 23.8 & 88.2 & 98.6 & [22.8, 32.8] \\
Retrosynthesis & 24.4 & 88.2 & 98.0 & [14.2, 21.4] \\
ChEBI-20 text to molecule & 36.1 & 79.0 & 80.8 & [13.5, 16.5] \\
 
\bottomrule
\end{tabular}
\end{table}

At the final refinement budget, retrosynthesis reaches 88.2\% trace soundness and 20.8\% exact set-match accuracy; ChEBI-20 reaches 79.0\% soundness and 33.1\% accuracy; and functional-group generation reaches 97.6\% acceptance and 41.8\% benchmark success. In these tasks, the declared obligations capture part of the instruction, and native functional-group patterns follow MolLean's own definitions.

For ChEBI-20, 1,075 outputs are both sound and correct, 1,533 are sound but incorrect, 17 are correct without certification, and 675 satisfy neither criterion. These exhaustive outcomes distinguish process acceptance from task success (Appendix~\ref{app:joint}). Atom-count generation shows the opposite profile: 444 of 467 accepted traces meet its task criterion, compared with 1,075 of 2,608 on ChEBI-20. A count maps directly to a predicate, whereas a textual description requires the model to select and combine the right structural obligations. Instruction-to-constraint grounding is therefore the main lever for further gains, as the salt example in Section~\ref{sec:cases} illustrates.

\subsection{Interactive Verification Strengthens Reaction Prediction}
Figure~\ref{fig:analysis}c shows the signed differences between the two interfaces. MCP achieves the best reaction results, with 66.4\% forward-reaction and 27.4\% retrosynthesis accuracy against 62.8\% and 20.8\% for Refinement. Refinement leads on atom counts (88.8\% versus 82.6\%), bond counts (69.0\% versus 56.2\%), QED optimization (59.0 versus 49.4 weighted-score points), and ChEBI-20 (33.1\% versus 23.5\%), and both protocols reach 100.0\% on component addition and 97.0\% on component deletion.

Whole-trace refinement can revise an entire plan, and MCP conditions each action on the maintained state and immediate feedback. Immediate feedback helps the model maintain references across a sequence of dependent edits, as in multi-step reactions, while restarting permits larger changes of plan. The two interfaces are therefore complementary. Appendix~\ref{app:mcp-calls} reports the interaction-length distributions.

\subsection{Whole-Trace Inference with Qwen3-4B}
The smaller-model evaluation shows substantial gains on several structural tasks (Table~\ref{tab:4b}). It uses the same task IDs and whole-trace protocols as the main evaluation.

\begin{table}[t]
\centering
\caption{\textbf{Qwen3-4B task performance.} The tasks and example IDs match the main evaluation. $\dagger$ denotes $100\times$ success-weighted similarity; other scores are success percentages.}
\label{tab:4b}
\small
\setlength{\tabcolsep}{6pt}
\begin{tabular}{lrrrr}
\toprule
Task & $N$ & CoT & One-pass & Refine \\
\midrule
Atom counts & 500 & 15.8 & 39.6 & 56.8 \\
Bond counts & 500 & 9.6 & 13.4 & 22.4 \\
Functional groups & 500 & 11.8 & 12.2 & 16.6 \\
Add component & 500 & 13.4 & 41.2 & 91.4 \\
Delete component & 500 & 59.0 & 56.4 & 64.2 \\
Substitute component & 500 & 15.2 & 21.6 & 53.4 \\
logP optimization $\dagger$ & 500 & 14.7 & 29.6 & 60.0 \\
MR optimization $\dagger$ & 500 & 19.4 & 15.3 & 32.0 \\
QED optimization $\dagger$ & 500 & 6.8 & 11.5 & 31.2 \\
Forward reaction & 500 & 1.4 & 2.2 & 1.6 \\
Retrosynthesis & 500 & 0.0 & 0.0 & 0.0 \\
ChEBI-20 text to molecule & 3,300 & 0.6 & 0.8 & 1.4 \\
 
\bottomrule
\end{tabular}
\end{table}
 Relative to Direct CoT, refinement raises atom-count success from 15.8\% to 56.8\%, component addition from 13.4\% to 91.4\%, substitution from 15.2\% to 53.4\%, and the logP optimization score from 14.7 to 60.0. Verification-guided refinement therefore transfers to a smaller model.

Reference-based generation remains difficult for the smaller model: after refinement, forward-reaction accuracy is 1.6\% (One-pass: 2.2\%), ChEBI-20 reaches 1.4\%, and retrosynthesis is 0.0\% under every protocol. These tasks depend on chemical knowledge that the proposer must supply, so verification is most effective when paired with a capable candidate generator.

\FloatBarrier
\section{Case Studies: Repairs and Specification Gaps}
\label{sec:cases}
\begin{figure}[!t]
\centering
\includegraphics[width=\linewidth]{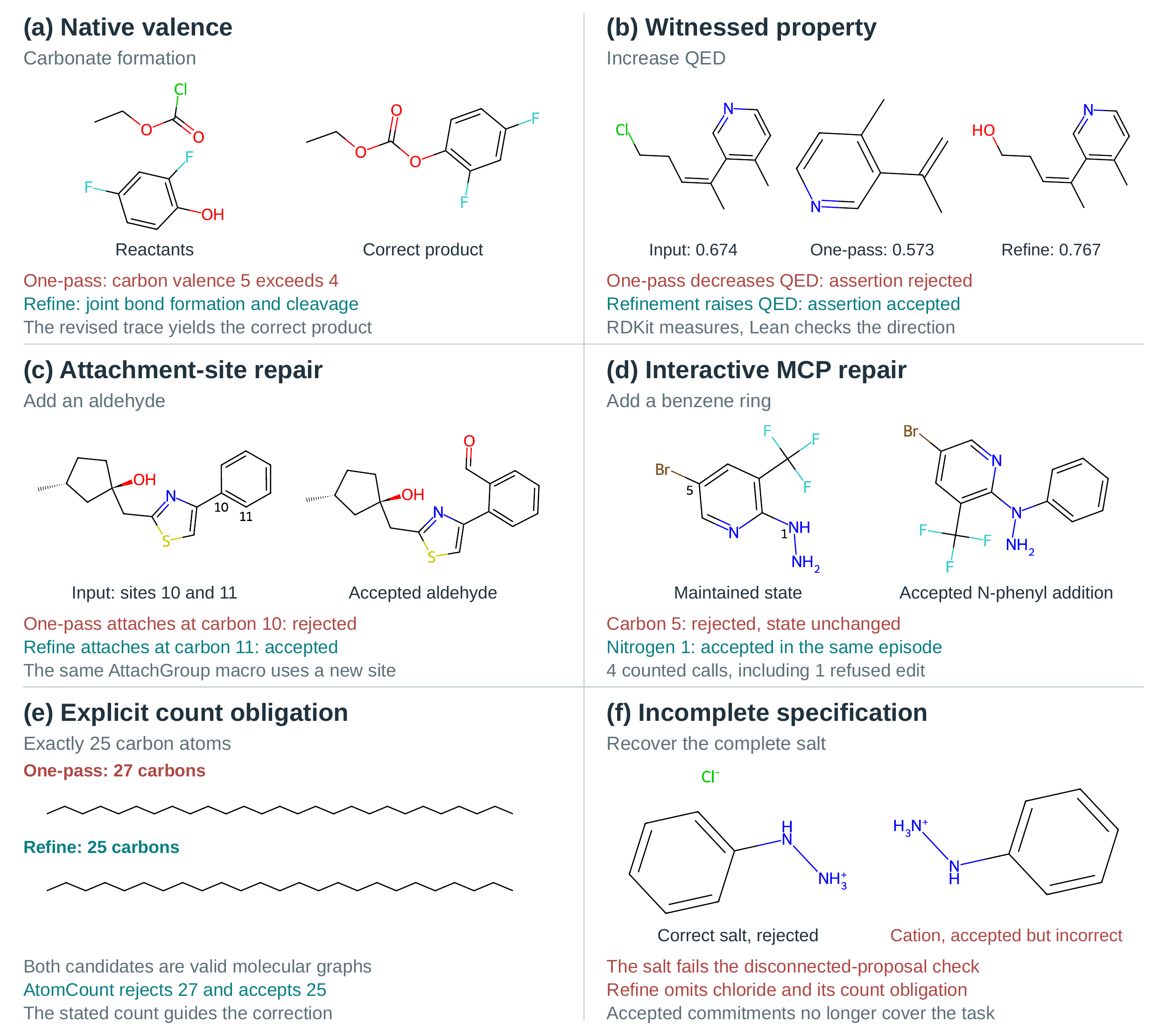}
\caption{\textbf{Molecular cases of repair and incomplete specification.} (a,c) Refinement changes action granularity or the attachment site. (b,e) A property or count assertion rejects an otherwise valid candidate. (d) MCP repairs an edit within the same maintained state. (f) A revised salt description drops a required component, giving an accepted but incorrect answer. Molecular drawings use the saved structures; numbers in (c,d) denote input atom indices. Refinement restarts from the task input, whereas MCP continues within its episode.}
\label{fig:cases}
\end{figure}

Figure~\ref{fig:cases} pairs molecular structures with verification feedback in six cases, illustrating how rejected commitments are repaired and how an accepted revision can still miss the intended task.

\paragraph{Valence and attachment sites.}
In the forward-reaction example (a), forming a bond to an acyl carbon before breaking its bond to chlorine would give carbon valence five. The one-pass trace is rejected. Refinement replaces the sequential edits with a joint bond-formation and bond-cleavage operation, removes the detached chlorine, and obtains the correct carbonate product. The aldehyde-addition example (c) instead needs a different attachment site: the same \op{AttachGroup} macro fails at substituted aromatic carbon 10 and succeeds at carbon 11. The revised trace then passes the functional-group-change and input-relative similarity checks. The carbonate requires suitable action granularity, whereas the aldehyde requires a suitable molecular referent.

\paragraph{Measured properties and explicit counts.}
In (b), the one-pass edits lower QED \citep{bickerton2012qed} from 0.674 to 0.573 despite an instruction to increase it. The graph edits pass, and the property assertion fails. Refinement reaches QED 0.767, with success-weighted similarity 0.743. RDKit supplies the measurement and Lean checks the requested direction. In (e), a valid 27-carbon chain fails \op{AtomCount} because the instruction requests 25. Refinement proposes a 25-carbon chain and passes that count. Both cases separate candidate validity from the requested property, with different evidence sources: atom count is native to the graph, and QED is an externally computed descriptor checked through a witness.

\paragraph{Interactive recovery and incomplete specifications.}
In (d), MCP's phenyl attachment at carbon 5 is refused without changing the state. The next edit attaches it to nitrogen 1, followed by two checks. The episode finishes correctly after four counted calls, including the refusal. In (f), one-pass proposes the correct salt and fails the disconnected-proposal check. Refinement omits chloride and its count obligation, producing an accepted cation that misses the target salt. Effective revision therefore requires both accurate execution and a specification that continues to express what the user asked for.

\FloatBarrier
\section{Related Work}
Chemical language research connects molecular representations with natural-language descriptions and instructions. Text2Mol introduced text-based molecular retrieval and ChEBI-20 \citep{edwards2021text2mol}; MolT5 studied translation between molecular and natural-language representations \citep{edwards2022molt5}; and Speak-to-Structure evaluates instruction-driven generation, editing, and optimization \citep{li2026speak}. SELFIES addresses molecular-string validity \citep{krenn2020selfies}, whereas structural reasoning targets molecular understanding \citep{jang2025structural}. ChemBench evaluates chemical knowledge and reasoning more broadly \citep{mirza2025chembench}. These perspectives complement evidence that fluent chemical reasoning can contain unreliable structural claims \citep{li2026scratchpad}. Among executable chemical languages, XDL expresses synthesis procedures for compilation into laboratory instructions \citep{mehr2020xdl}. 

Formal and executable reasoning make different parts of a derivation checkable. Lean separates proof construction from proof checking \citep{demoura2021lean4}; Draft, Sketch, and Prove links informal planning to formal proofs \citep{jiang2023draft}, and LeanDojo supports programmatic interaction with Lean \citep{yang2023leandojo}. Learned process supervision evaluates intermediate reasoning with a trained verifier \citep{lightman2024verify}. PAL and Program of Thoughts use execution to ground generated computations \citep{gao2023pal,chen2023pot}. 

Feedback and tool use turn evaluation into part of generation. Self-Refine uses model-generated feedback \citep{madaan2023selfrefine}, and CRITIC grounds revision in external tools \citep{gou2024critic}, which is more reliable than intrinsic self-correction on reasoning tasks \citep{huang2024selfcorrect}. ReAct interleaves reasoning and actions \citep{yao2023react}. In chemistry, ChemCrow integrates specialized tools \citep{bran2024chemcrow}, Coscientist connects planning with experimental automation \citep{boiko2023coscientist}, and ChemReasoner steers catalyst search using computational-chemistry feedback \citep{sprueill2024chemreasoner}. 

\section{Conclusion}
MolLean grounds chemical reasoning in a maintained molecular state, combining Lean-checked obligations with explicit external witnesses. MolLean-MCP makes this feedback available during generation. Whole-trace refinement improves Qwen3.6-27B performance across all 12 tasks, MCP delivers the strongest reaction results, and Qwen3-4B gains substantially on structural tasks. Grounding instructions in explicit constraints is the natural next step toward chemical reasoning that is both verifiable and faithful to the task.

\label{main-text-end}
\clearpage
\label{references-start}
\bibliography{references}
\bibliographystyle{iclr2025_conference}
\clearpage
\appendix
\raggedbottom
\section{Implementation Details}
\label{app:implementation}
MolLean combines Lean checks with Python orchestration and external evidence providers. Table~\ref{tab:implementation} distinguishes their responsibilities.

\begin{table}[htbp]
\centering
\caption{\textbf{Verification components and execution semantics.}}
\label{tab:implementation}
\small
\begin{tabular}{>{\raggedright\arraybackslash}p{0.22\linewidth}>{\raggedright\arraybackslash}p{0.35\linewidth}>{\raggedright\arraybackslash}p{0.34\linewidth}}
\toprule
Component & Implemented check & Execution semantics \\
\midrule
Molecule validator & Proved valence and connectivity soundness; conditional aromaticity & Validation of represented molecular graphs \\
Native assertions & Graph predicates; six assertion-boundary theorems & Checks on explicitly stated obligations \\
Edits and bundles & Preconditions, valences, bundle outcomes & Atomic validation of a concerted bundle's outcome \\
Bookkeeping & Atoms, charge, radical parity across both pools & Additions and removals update the ledger \\
External witnesses & Typed request and predicate on returned fields & Provider and witness binding trusted \\
Python orchestration & Parsing, reference translation, serialization, control flow & State transport and inference control \\
\bottomrule
\end{tabular}
\end{table}

The validator's \op{validate\_sound} theorem concerns the returned proof-carrying molecular object. Native assertion-boundary proofs cover \op{AtomCount}, \op{BondCount}, \op{RingCount}, \op{HasAromatic}, \op{FGCount}, and \op{FormulaMatch}. Hydrogen counts include graph and implicit hydrogens. Formula matching covers every present element and rejects duplicate element entries, while bookkeeping tracks literal graph atoms.

\definecolor{appendixink}{HTML}{5E6C82}
\definecolor{appendixrule}{HTML}{B7CBD5}
\definecolor{appendixpaper}{HTML}{F7F9F9}
\lstdefinestyle{appendixcode}{basicstyle=\footnotesize\ttfamily,
  columns=fullflexible,keepspaces=true,breaklines=true,breakatwhitespace=false,
  showstringspaces=false,upquote=false,frame=single,framerule=0.3pt,
  rulecolor=\color{appendixrule},backgroundcolor=\color{appendixpaper},
  xleftmargin=0.6em,xrightmargin=0.3em,framexleftmargin=0.4em,
  aboveskip=0.5em,belowskip=0.5em,captionpos=t,
  postbreak=\mbox{\textcolor{appendixink}{$\hookrightarrow$}\space}}
\lstdefinestyle{appendixprompt}{style=appendixcode,
  basicstyle=\footnotesize\ttfamily,breakatwhitespace=false}
\newcommand{\fixedlisting}[2][]{\par\noindent
  \begin{minipage}{\linewidth}\lstinputlisting[#1]{#2}\end{minipage}\par}

\subsection{Represented State and Reference Resolution}
The runtime serializes a \op{MultiMolState} between Python and a persistent Lean process. Its core fields are reproduced below; each \op{TaggedMol} pairs a molecule identifier with a graph.
\begin{lstlisting}[style=appendixcode,caption={Core Lean state declaration (comments omitted).},label={lst:state}]
structure MultiMolState where
  active              : List TaggedMol
  leaving_pool        : List TaggedMol
  step_count          : Nat
  init_atom_count     : Nat
  init_charge         : Int
  init_radical_parity : Nat

structure AtomRef where
  mol_id   : MolId
  atom_idx : AtomIdx
\end{lstlisting}
A molecular graph stores indexed atoms, bonds, and stereochemical annotations. Atom fields include element, formal charge, radicals, aromaticity, and virtual hydrogen information; bond fields specify endpoints and order. Virtual hydrogens contribute to element and formula readouts, and the bookkeeping ledger counts graph atoms. The \op{leaving\_pool} is excluded from the predicted answer and remains part of that ledger. Additions, removals, and proposals update the atom-count ledger.

References are pairs \op{(mol\_id, atom\_idx)} into the graph state. A cross-molecule bond joins two index spaces, and a fragment split creates a new molecule identifier. In whole-trace execution, Python tracks accepted primitive merges and remaps subsequent step references through that history. Lean separately rebases references inside tracked compound bundles. MCP uses the current atom map returned in feedback for later model calls. Python also exports the maintained graphs to molecular strings.

\subsection{Action Dispatch and Commit Semantics}
Complete traces are JSON objects with a \op{trace} array. A step identifies an operation family, for example \op{op} with a primitive, \op{macro} with a construction macro, \op{compound} with a joint transformation, or \op{assert} with a claim. SMILES proposals are parsed into graphs before validation. MCP functions translate to these same step representations and run on the same chemical kernel.

Listing~\ref{lst:dispatch} summarizes the dispatch path. Lean returns either an accepted state or a structured error. The runtime replaces the maintained state only after acceptance. A failed operation can therefore report an attempted edit without adding it to the accepted trace. Native assertions and external comparisons inspect the molecular graph without changing it.
\begin{lstlisting}[style=appendixcode,caption={Stateful dispatch and the external-evidence boundary (pseudocode).},label={lst:dispatch}]
action = translate(model_call, current_state)
action = discard_model_supplied_witness(action)
reply = lean.apply_action(current_state, action)

if reply.need_witness:
    witness = provider.answer(reply.request, initial_input)
    reply = lean.apply_action(current_state,
                              action.with_witness(witness))

if reply.accept:
    current_state = reply.state
    record_accepted_action(action)
    return current_state_and_indices(current_state)
else:
    return rejection_and_unchanged_state(reply.error, current_state)
\end{lstlisting}
Primitive edits validate their preconditions and affected valences. A concerted bundle validates its joint outcome, which accommodates edits whose sequential ordering would pass through a temporarily over-valent graph. The closing procedure checks charge and radical bookkeeping and the task's terminal obligations. Realizability and single-output requirements are task dependent (Appendix~\ref{app:tasks}).

\subsection{A Concrete Typed Witness}
Native functional-group checks use Lean's built-in patterns. A pattern without a native definition requests an RDKit \op{fg\_count} witness, whose returned count Lean compares with the assertion. Native \op{StereoCenterCount} counts retained R/S annotations. Bonding changes at a tracked center conservatively remove its annotation.

For native \op{AtomCount}, Lean counts the represented atoms of the requested element, including the applicable hydrogen convention; no provider-supplied count is needed. In contrast, \op{PropertyChange(qed)} requests an RDKit measurement. The provider evaluates the current molecule and the bound initial input, then supplies integer-valued fields. Lean requires \op{computable=1} and compares \op{out\_x1000} with \op{in\_x1000} in the requested direction.

Listing~\ref{lst:witness} is the actual accepted payload from the QED case in Appendix~\ref{app:case-qed}. Its values encode an increase from approximately $0.674$ to $0.767$. Lean checks the predicate on these returned typed fields.
\fixedlisting[style=appendixcode,caption={Accepted QED assertion and its supplied witness.},label={lst:witness}]{listings/implementation/qed_witness.json}

\FloatBarrier
\Needspace{12\baselineskip}
\section{Prompts and Tool Interface}
\label{app:prompts}
This section documents the prompts used by the evaluation interfaces. Prompt text is reproduced from the implemented templates or a saved MCP conversation. Line wrapping and typographic glyphs are normalized for typesetting; task variables retain their actual values.

\subsection{Direct CoT and Whole-Trace Generation}
Direct CoT combines Listing~\ref{lst:cot-system} with the dataset task instruction. No verifier state, feedback, or held-out answer is included. One-pass uses Listing~\ref{lst:trace-system} and a task-specific user message containing the instruction, any loaded molecular input and atom map, operator vocabulary, and JSON output contract. Refinement shares that initial completion and prompt.
The whole-trace prompt's phrase ``each step is checked'' refers to sequential execution of a completed trace: the model generates that trace before receiving any rejection feedback.
\begin{lstlisting}[style=appendixprompt,caption={Direct CoT system prompt.},label={lst:cot-system}]
You are an expert chemist. Reason step by step about the task, then 
output your final answer on the LAST line as 'ANSWER: <SMILES>' 
-- just the SMILES, nothing else.
\end{lstlisting}
\begin{lstlisting}[style=appendixprompt,caption={Whole-trace system prompt shared by One-pass and Refinement.},label={lst:trace-system}]
You are an expert chemist whose every reasoning
step is verified by a Lean 4 theorem prover (MolLean). Emit a chain
of typed reasoning steps that build up to (or describe) the target
molecule. Each step is checked by Lean before the next is allowed.
A single rejected step rejects the whole chain.

Output strict JSON only -- no prose outside the JSON block.
\end{lstlisting}

Listing~\ref{lst:generation-prompt} instantiates the atom-count generation template for a real example. The marked insertion point expands to the shared vocabulary in Listing~\ref{lst:vocabulary}, which the model receives in full. The complete instantiated user messages for all 12 tasks accompany the LaTeX source as \op{listings/prompts/full\_\textit{task}.txt}.
\begin{lstlisting}[style=appendixprompt,caption={Atom-count generation user message, with shared vocabulary factored out.},label={lst:generation-prompt}]
Task: mol_custom_atomnum  (generate a molecule satisfying a spec)

Spec:
There is a molecule with 25 carbon atoms.

Build a molecule satisfying the spec. Pick the easiest route:

  1. FASTEST: `propose` a complete candidate SMILES that meets the spec
     (MolLean validates it), or `propose` a core scaffold and then edit.
     You may also propose several fragments and join them with a
     cross-molecule BondForm.
  2. Or build it up: Chain / Ring macros for the skeleton (a chain of N
     carbons is ONE Chain macro), AttachGroup / AttachRing for
     substituents; individual Op primitives for fine adjustments.
  3. Assert steps restating every numerical constraint from the spec
     (AtomCount per element, BondCount per type, FGCount per group,
     RingCount, HasAromatic).

NOTE: the FINAL molecule is checked for physical realizability --
valence-saturatable atoms, single connected component, Huckel-valid
aromatic rings, no leftover unpaired electrons. Finish on a real 
molecule.

[Insert the shared operator vocabulary here.]

Return JSON:
{ "strategy":"...", "trace":[<step list>] }
\end{lstlisting}

\begin{table}[htbp]
\centering
\caption{\textbf{Task-specific prompt content.} All whole-trace templates append the same operator vocabulary. Reference structures for editing and optimization are the input molecule, not a held-out answer.}
\label{tab:prompt-families}
\small
\begin{tabular}{>{\raggedright\arraybackslash}p{0.23\linewidth}>{\raggedright\arraybackslash}p{0.34\linewidth}>{\raggedright\arraybackslash}p{0.34\linewidth}}
\toprule
Family & Supplied input & Requested construction and checks \\
\midrule
Atom, bond, and group generation & Natural-language specification; empty state & Construct or propose; assert requested counts \\
Add, delete, and substitute & Input SMILES and indexed atom/bond map & Edit in place; group-count change and input similarity \\
logP, MR, and QED optimization & Input molecule and requested direction & Property-changing edits; direction and input similarity \\
Forward reaction & Reactants, reaction class, and atom maps & Transform loaded reactants; check proposed outcome \\
Retrosynthesis & Product, reaction class, and atom map & Disconnect and modify fragments; check proposed outcome \\
Text to molecule & Molecular description; empty state & Propose or construct; assert explicit structural requirements \\
\bottomrule
\end{tabular}
\end{table}

\subsection{Whole-Trace Revision Feedback}
After a rejection, the feedback identifies the first failed step, its error, and the state at that point. When available, it includes the live atom map. Listing~\ref{lst:repair-prompt} shows the full feedback produced for the acylation case. This is a request for a corrected \emph{complete} trace, which is executed from the original input. No correctness score or held-out target enters this message.
\begin{lstlisting}[style=appendixprompt,caption={Whole-trace revision feedback for a rejected acyl-carbon edit.},label={lst:repair-prompt}]
MolLean (the Lean verifier) REJECTED your reasoning trace.
Reason: step 0 rejected: atom (mol=0, idx=3) would have effective 
valence x2 (bond orders + radicals) 10; charge-adjusted allowed x2: [8]
The FIRST rejected step was index 0 (op): atom (mol=0, idx=3) would have
effective valence x2 (bond orders + radicals) 10; charge-adjusted 
allowed x2: [8]
State just before it failed: CCOC(=O)Cl, Oc1ccc(F)cc1F
LIVE ATOM MAP at that point -- use EXACTLY these (mol_id, atom_idx)
  coordinates in your fix:
mol 0 ('CCOC(=O)Cl'): atoms [0]C [1]C [2]O [3]C [4]O [5]Cl | bonds 0-1 
1-2 2-3 3=4 3-5
mol 1 ('Oc1ccc(F)cc1F'): atoms [0]O [1]C [2]C [3]C [4]C [5]F [6]C [7]C 
[8]F | bonds 0-1 1:2 2:3 3:4 4-5 4:6 6:7 7-8 7:1

Diagnose this specific failure and return the FULL corrected trace as 
JSON (same schema: {"strategy": "...", "trace": [ ... ]}). Keep the 
steps that were already accepted; fix the failing step and everything 
downstream of it.
Do not repeat the same mistake.
\end{lstlisting}
An unparseable shared initial completion ends both One-pass and Refinement unsuccessfully. An unparseable later revision stops refinement at the last parseable outcome; only executed revisions enter the round counter.

\subsection{Stateful MCP Prompt and Schemas}
MCP supplies function schemas to the model and maintains the molecular state across calls. Listing~\ref{lst:mcp-system} reproduces its system brief. Listing~\ref{lst:mcp-user} is the complete initial user message from the phenyl-attachment episode in Appendix~\ref{app:case-mcp}. It contains the task instruction and current atom map. Later turns append the actual verifier response; a rejected edit leaves this maintained state unchanged.
\begin{lstlisting}[style=appendixprompt,caption={MCP system brief.},label={lst:mcp-system}]
You are deriving chemistry with a verifier attached, and the verifier is
fast. Do not analyse the whole route before acting. Decide the most 
likely next edit, call the function, and read what comes back: 
the kernel answers in milliseconds and its reply is better evidence than 
any amount of reasoning about what the reply would be.

Every reply must end in exactly one function call. One or two sentences 
of reasoning before it is enough; a paragraph is a paragraph the kernel 
cannot check. If you are unsure which of two edits is right, try one. 
A refused call costs a call and leaves the state untouched, and the 
refusal names the reason.

Atom indices always come from the kernel's last reply. Never count them 
off a SMILES string. Call finish() when the state holds the answer.
\end{lstlisting}
\begin{lstlisting}[style=appendixprompt,caption={Saved initial MCP user message for phenyl attachment.},label={lst:mcp-user}]
Please add a benzene ring to the molecule NNc1ncc(Br)cc1C(F)(F)F.

Current state:
mol 0 ('NNc1ncc(Br)cc1C(F)(F)F'): atoms [0]N [1]N [2]C [3]N [4]C [5]C 
[6]Br [7]C [8]C [9]C [10]F [11]F [12]F | bonds 0-1 1-2 2:3 3:4 4:5 5-6 
5:7 7:8 8-9 9-10 9-11 9-12 8:2
\end{lstlisting}

The schemas distinguish action arguments from measured evidence. For example, \op{attach\_group} names a molecule, anchor atom, and group; \op{check} names a claim and its JSON arguments; \op{finish} requests the closing procedure. Descriptor values always come from a provider. Listing~\ref{lst:tool-schema} shows the complete attachment schema. The full schema inventory is included in the source file \op{listings/prompts/all\_tools.json}, together with separate \op{check} and \op{finish} schemas.
\fixedlisting[style=appendixcode,caption={Native function schema for attaching a group.},label={lst:tool-schema}]{listings/prompts/tool_attach_group.json}

\subsection{Shared Whole-Trace Vocabulary}
Listing~\ref{lst:vocabulary} reproduces the shared operator instructions inserted into the task-specific whole-trace messages. These instructions describe the model-facing syntax; task setup and runtime checks determine which proposals and closing conditions apply (Appendix~\ref{app:tasks}). In particular, the runtime permits candidate proposals in optimization and reactions, and requires a single final molecule only for editing. The vocabulary's atom-count shorthand for new indices assumes contiguous indexing; the implementation allocates after the largest current atom index.
\begin{lstlisting}[style=appendixprompt,caption={Shared whole-trace operator vocabulary.},label={lst:vocabulary}]
Available step shapes:

  -- Propose steps -- TWO distinct interfaces --------------------------
  MolLean validates every proposed structure (no impossible valences,
  single connected component) before it enters the state.

    {"step_type":"propose","smiles":"<SMILES>"}
        -- Propose a WHOLE molecule. Allowed only for SYNTHESIS tasks
          (text_to_mol / mol_custom), where you build a target from
          nothing. If the state is empty it becomes mol_id 0.

    {"step_type":"propose_fragment","smiles":"<SMILES>"}
        -- Propose a FRAGMENT to graft onto an existing molecule. It is
          added as a new mol_id; you THEN attach it with a 
          cross-molecule BondForm (src in mol 0, dst in the fragment), 
          which merges it into mol 0. Use this for TRANSFORMATION tasks 
          (mol_edit / mol_opt) -- e.g. enumerate property-improving 
          substructures and combine them onto the input. The result must 
          end as ONE molecule.

  Idioms: base-and-edit (propose a scaffold, then AttachGroup/BondForm/
  AddAtom/RemoveAtom); fragment assembly (propose_fragment several 
  pieces, join with cross-molecule BondForm).

  -- Op steps (state-changing primitives) --------------------------
    {"step_type":"op","primitive":{"type":"AddAtom","mol_id":0,
        "element":"C"|"N"|"O"|...,"charge":0,"radicals":0,
        "is_aromatic":false}}
        -- Append a new atom to molecule `mol_id`. Atom indices grow
          sequentially as you add them (first AddAtom -> idx 0).
    {"step_type":"op","primitive":{"type":"BondForm",
        "src":{"mol_id":0,"atom_idx":i},
        "dst":{"mol_id":0,"atom_idx":j},
        "bond_type":"Single"|"Double"|"Triple"|"Aromatic"}}
        -- Form a bond between two existing atoms.
    {"step_type":"op","primitive":{"type":"BondBreak",
        "src":{"mol_id":0,"atom_idx":i},
        "dst":{"mol_id":0,"atom_idx":j}}}
    {"step_type":"op","primitive":{"type":"BondOrderChange",
        "src":{...},"dst":{...},"bond_type":"..."}}
    {"step_type":"op","primitive":{"type":"ChargeUpdate",
        "atom":{"mol_id":0,"atom_idx":i},"delta":+1|-1}}
    {"step_type":"op","primitive":{"type":"RadicalUpdate",
        "atom":{...},"delta":+1|-1}}
    {"step_type":"op","primitive":{"type":"RemoveAtom",
        "atom":{"mol_id":0,"atom_idx":i}}}

  -- Compound steps (SUBGRAPH-LEVEL mechanism ops) --------------------
  Act on a GROUP of atoms/bonds at once -- useful for reaction 
  mechanisms.
  Each desugars to Layer-1 primitives that Lean verifies; conservation 
  is checked over the whole step.
    {"step_type":"compound","compound":{"compound":"ProtonShift",
        "h":{...},"donor":{...},"acceptor":{...}}}
        -- move a proton H from donor to acceptor (acid/base, tautomer).
          The state is HEAVY-ATOM only: implicit Hs are not addressable.
          First AddAtom an explicit H + BondForm it to the donor, then
          ProtonShift that H. Works across molecules (the framework
          re-bases refs through the merge).
    {"step_type":"compound","compound":{"compound":"Eject",
        "anchor":{...},"leaving":{...}}}
        -- break anchor-leaving and send the leaving group to the pool.
    {"step_type":"compound","compound":{"compound":"Fragment",
        "anchor":{...},"leaving":{...}}}
        -- break a BRIDGE bond and split the molecule into TWO active
          molecules (retro: product -> two reactant fragments). Both 
          halves stay active.
    {"step_type":"compound","compound":{"compound":"BondFormBreak",
        "form_src":{...},"form_dst":{...},"form_bt":"Single",
        "break_src":{...},"break_dst":{...}}}
        -- concerted form-one-bond + break-another (SN2, addition-elim).
    {"step_type":"compound","compound":{"compound":"Concerted",
        "primitives":[ <op primitive>, ... ]}}
        -- apply a bundle of primitives atomically (Lean checks the net
          result).

  -- Macro steps (HIGH-LEVEL build ops -- STRONGLY PREFERRED) -----------
  Prefer these over long AddAtom/BondForm chains. Each macro expands to
  Layer-1 primitives that Lean still verifies individually, but your
  reasoning stays compact and you avoid atom-index bookkeeping errors.
  After a macro, atom indices continue from where the macro left off
  (a Chain/Ring of n atoms occupies the next n free indices, in order).

    {"step_type":"macro","macro":{"macro":"Chain",
        "element":"C","n":N,"bond_type":"Single","mol_id":0}}
        -- A linear chain of N atoms. Occupies indices [base .. base+N-1] where base = current atom count.
    {"step_type":"macro","macro":{"macro":"Ring",
        "element":"C","n":N,"aromatic":true,"mol_id":0}}
        -- An N-membered ring (aromatic if "aromatic":true -> benzene for N=6). Occupies the next N indices and closes the loop.
    {"step_type":"macro","macro":{"macro":"AttachGroup",
        "anchor":{"mol_id":0,"atom_idx":i},"group":"<name>"}}
        -- Graft a functional group onto existing atom i. Recognised
          groups: hydroxyl, oxo, amino, methyl, carboxyl, carboxylate,
          aldehyde, ester, amide, nitro, nitrile, methoxy, ether,
          sulfonyl, phosphate, fluoro/chloro/bromo/iodo, thiol,
          benzene/phenyl, pyridine, furan, pyrrole, thiophene,
          cyclohexane, cyclopropane, cyclopentane.  (Or pass a raw
          fragment SMILES as "group".)
    {"step_type":"macro","macro":{"macro":"AttachRing",
        "anchor":{"mol_id":0,"atom_idx":i},"n":6,"aromatic":true,
        "mol_id":0}}
        -- Build an N-membered ring and bond it to existing atom i.

  -- Assert steps (state-querying claims; witness filled by framework) --
    {"step_type":"assert","assertion":{"claim":"AtomCount",
        "element":"carbon"|"nitrogen"|...,"count":N,"mol_id":0}}
    {"step_type":"assert","assertion":{"claim":"BondCount",
        "bond_type":"Single"|"Double"|...,"count":N,"mol_id":0}}
{"step_type":"assert","assertion":{"claim":"RingCount","count":N,"mol_id":0}}
{"step_type":"assert","assertion":{"claim":"HasAromatic","expected":true,"mol_id":0}}
    {"step_type":"assert","assertion":{"claim":"FGCount",
        "fg":"<recognised-name>","count":N,"mol_id":0}}
    {"step_type":"assert","assertion":{"claim":"FGCountDelta",
        "fg":"<name>","delta":+1|-1,"input_smiles":"...","mol_id":0}}
{"step_type":"assert","assertion":{"claim":"NameMatch","name":"...","mol_id":0}}
{"step_type":"assert","assertion":{"claim":"ClassMember","class":"...","mol_id":0}}
    {"step_type":"assert","assertion":{"claim":"PropertyChange",
"prop":"logp"|"qed"|"solubility"|...,"direction":"increase"|"decrease",
        "input_smiles":"...","mol_id":0}}
    {"step_type":"assert","assertion":{"claim":"OutcomeMatch",
        "reference_smiles":"<target>","mode":"skeleton"|"exact"|"tanimoto", "mol_id":0}}
{"step_type":"assert","assertion":{"claim":"StereoCenterCount","count":N,"mol_id":0}}
        -- exactly N tracked R/S stereocentres. Conservative: an edit AT 
        a centre drops its annotation, so only claim centres your edits
        never touched.

Recognised FG names (for FGCount / FGCountDelta):
hydroxyl, aldehyde, ketone, carboxyl, carboxylate, ester, amide,
amine, primary_amine, secondary_amine, tertiary_amine, nitro,
nitrile, thiol, thioether, sulfoxide, sulfone, halide, fluoride,
chloride, bromide, iodide, alkene, alkyne, ether, epoxide, imine,
phenol, anhydride, benzene_ring, pyridine, pyrrole, furan,
thiophene, indole, imidazole, naphthalene, quinoline, pyrimidine,
phosphate_ester, sulfate_ester.
\end{lstlisting}

\FloatBarrier
\Needspace{12\baselineskip}
\section{Worked Examples}
\label{app:examples}
These eight examples show the actual submitted operations, verifier feedback, and saved outcomes. The first six expand the cases in Figure~\ref{fig:cases}; the last two illustrate a multi-molecule retrosynthetic answer and a native functional-group assertion. JSON is reformatted without changing fields or values. Whole-trace listings retain every submitted step, including unexecuted steps after the first rejection. MCP listings retain every function call and verifier reply; the model's free-form deliberation is omitted for space.

\Needspace{12\baselineskip}
\subsection{Acylation: Changing the Verification Unit}
\label{app:case-acylation}
\begin{lstlisting}[style=appendixprompt,caption={Forward-reaction task input.}]
Predict the major product(s) of the following acylation reaction.
Reactants: CCOC(=O)Cl.Oc1ccc(F)cc1F
Output only the product SMILES. Use a dot to separate multiple products.
\end{lstlisting}
The first trace forms the carbonyl-carbon--phenolic-oxygen bond before removing the carbon--chlorine bond. The first edit is rejected: the acyl carbon would have bond-order valence five. In the diagnostic representation, valence is doubled, so the attempted value 10 exceeds the allowed 8. The subsequent ejection and proposed-outcome assertion remain unexecuted.
\fixedlisting[style=appendixcode,caption={Complete initial acylation trace; execution stops at step 0.}]{listings/cases/acylation_B.json}
\begin{lstlisting}[style=appendixprompt,caption={First-rejection feedback.}]
step 0 rejected: atom (mol=0, idx=3) would have effective valence x2 
(bond orders + radicals) 10; charge-adjusted allowed x2: [8]
\end{lstlisting}
The corrected trace uses \op{BondFormBreak} to validate bond formation and cleavage as one transformation, then removes the detached chlorine atom. Its three steps are accepted and its final carbonate product is correct. The \op{OutcomeMatch} structure is the model's own proposed endpoint. The change repairs a graph-execution failure.
\fixedlisting[style=appendixcode,caption={Complete accepted acylation revision, executed from the reactants.}]{listings/cases/acylation_C.json}

\Needspace{12\baselineskip}
\subsection{QED Optimization: A Valid Edit Can Move the Wrong Way}
\label{app:case-qed}
\begin{lstlisting}[style=appendixprompt,caption={QED task input.}]
Please optimize the molecule C/C(=C/CCCl)c1cnccc1C to have a higher QED 
value.
\end{lstlisting}
The initial trace deletes three atoms. These edits are individually accepted, and the subsequent \op{PropertyChange} assertion is rejected because QED decreases from $0.674$ to $0.573$. The final input-similarity assertion remains unexecuted. The feedback targets the stated objective.
\fixedlisting[style=appendixcode,caption={Complete initial QED trace; the property assertion fails.}]{listings/cases/qed_B.json}
\begin{lstlisting}[style=appendixprompt,caption={Measured-property rejection.}]
step 3 rejected: assertion PropertyChange(qed) does not hold: property 
'qed' did not increase (in 674/1000, out 573/1000)
\end{lstlisting}
The revision removes chlorine and attaches hydroxyl at atom 4. All four steps are accepted; the saved final structure is \lstinline[style=appendixcode]!C/C(=C/CCO)c1cnccc1C!, with QED approximately $0.767$. Appendix~\ref{app:implementation} shows the associated typed witness.
\fixedlisting[style=appendixcode,caption={Complete accepted QED revision.}]{listings/cases/qed_C.json}

\Needspace{12\baselineskip}
\subsection{Aldehyde Attachment: Repairing the Referent}
\label{app:case-aldehyde}
\begin{lstlisting}[style=appendixprompt,caption={Aldehyde-addition task input.}]
Modify the molecule C[C@@H]1CC[C@](O)(Cc2nc(-c3ccccc3)cs2)C1 by adding a
aldehyde.
\end{lstlisting}
Attaching aldehyde at atom 10 is rejected because that substituted aromatic carbon cannot accept the additional bond. The revision retains the same \op{AttachGroup} macro and moves its anchor to atom 11. Its group-count-change and input-similarity assertions then pass. This repair changes only the molecular referent.
\fixedlisting[style=appendixcode,caption={Initial aldehyde trace with the invalid attachment site.}]{listings/cases/aldehyde_B.json}
\begin{lstlisting}[style=appendixprompt,caption={Attachment-site rejection.}]
step 0 rejected: atom (mol=0, idx=10) would have effective valence x2 
(bond orders + radicals) 10; charge-adjusted allowed x2: [8]
\end{lstlisting}
\fixedlisting[style=appendixcode,caption={Complete aldehyde revision with anchor 11.}]{listings/cases/aldehyde_C.json}

\Needspace{12\baselineskip}
\subsection{MCP Phenyl Attachment: Continuing After a Refusal}
\label{app:case-mcp}
The task and initial indexed state are reproduced in Listing~\ref{lst:mcp-user}. This episode continues from the maintained state after the rejection. The first \op{attach\_group} call targets carbon 5 and fails. The reply explicitly reports \op{state\_unchanged=true} and the same atom map. The model next targets nitrogen 1; the accepted attachment introduces indices 13--18. Two checks follow: a benzene-ring count increase and similarity to the input.

Listing~\ref{lst:mcp-episode} shows all five function calls, including \op{finish}, and all four intervening verifier replies. The episode records four \emph{counted} calls, because successful \op{finish} is outside the edit/check counter. One of those four calls is refused, and the final answer is correct. The \op{OutcomeMatch} check is a Tanimoto comparison.
\begin{lstlisting}[style=appendixcode,caption={Saved MCP episode, with deliberation omitted and calls normalized to JSON.},label={lst:mcp-episode}]
MODEL CALL
{"name":"attach_group","arguments":{"mol_id":0,"atom_idx":5,"group":"phenyl"}}

VERIFIER REPLY
{
  "ok":false,
  "reason":"atom (mol=0, idx=5) would have effective valence x2 (bond orders + radicals) 10; charge-adjusted allowed x2: [8]",
  "state_unchanged":true,
  "indices":"mol 0 ('NNc1ncc(Br)cc1C(F)(F)F'): atoms [0]N [1]N [2]C [3]N [4]C [5]C [6]Br [7]C [8]C [9]C [10]F [11]F [12]F | bonds 0-1 1-2 2:3 3:4 4:5 5-6 5:7 7:8 8-9 9-10 9-11 9-12 8:2"
}

MODEL CALL
{"name":"attach_group","arguments":{"atom_idx":1,"group":"phenyl","mol_id":0}}

VERIFIER REPLY
{
  "ok":true,
  "molecules":["NN(c1ccccc1)c1ncc(Br)cc1C(F)(F)F"],
  "changed":"mol 0: new atom index 13, 14, 15, 16, 17, 18"
}

MODEL CALL
{
  "name":"check",
  "arguments":{
    "arguments":{"fg":"benzene_ring","delta":1,"input_smiles":"NNc1ncc(Br)cc1C(F)(F)F"},
    "claim":"FGCountDelta",
    "mol_id":0
  }
}

VERIFIER REPLY
{"ok":true,"molecules":["NN(c1ccccc1)c1ncc(Br)cc1C(F)(F)F"],"changed":"indices unchanged"}

MODEL CALL
{
  "name":"check",
  "arguments":{
    "arguments":{"reference_smiles":"NNc1ncc(Br)cc1C(F)(F)F","mode":"tanimoto"},
    "claim":"OutcomeMatch",
    "mol_id":0
  }
}

VERIFIER REPLY
{"ok":true,"molecules":["NN(c1ccccc1)c1ncc(Br)cc1C(F)(F)F"],"changed":"indices unchanged"}

MODEL CALL
{"name":"finish","arguments":{}}
\end{lstlisting}

\Needspace{12\baselineskip}
\subsection{Atom Count: The Count Obligation Guides the Correction}
\label{app:case-count}
\begin{lstlisting}[style=appendixprompt,caption={Atom-count task input.}]
There is a molecule with 25 carbon atoms.
\end{lstlisting}
The first proposal is a graph-valid chain with 27 carbon atoms. Its explicit \op{AtomCount} assertion requests 25 and is rejected by native counting. The corrected trace proposes a 25-carbon chain and repeats the same obligation. Both steps of the revision are accepted, and the generated molecule satisfies the requested count. No external measurement is needed for this rejection or repair.
\fixedlisting[style=appendixcode,caption={Initial 27-carbon proposal and its 25-carbon obligation.}]{listings/cases/atom_count_B.json}
\begin{lstlisting}[style=appendixprompt,caption={Native atom-count rejection.}]
step 1 rejected: assertion AtomCount(C) does not hold: 
expected 25 atoms, Lean counted 27
\end{lstlisting}
\fixedlisting[style=appendixcode,caption={Complete accepted 25-carbon revision.}]{listings/cases/atom_count_C.json}

\Needspace{12\baselineskip}
\subsection{Hydrochloride Description: Acceptance After Dropping a Requirement}
\label{app:case-salt}
\begin{lstlisting}[style=appendixprompt,caption={Text-to-molecule task input.}]
Given the following molecule description, output a SMILES string that
satisfies it.
Description: The molecule is a hydrochloride resulting from the reaction 
of equimolar amounts of phenylhydrazine and hydrogen chloride. It 
contains a phenylhydrazine.
\end{lstlisting}
The first completion proposes phenylhydrazinium chloride and explicitly requires one chlorine atom. Its disconnected proposal is rejected at step 0, before the assertions run. The revision proposes only the cation and removes both chloride and the chlorine-count obligation. Its four steps are accepted, and the final molecule misses the hydrochloride salt described in the task.
\fixedlisting[style=appendixcode,caption={Initial salt proposal and all submitted count obligations.}]{listings/cases/salt_B.json}
\begin{lstlisting}[style=appendixprompt,caption={Disconnected-proposal rejection.}]
step 0 rejected: this step would leave atom (mol=0, idx=0) disconnected
\end{lstlisting}
\fixedlisting[style=appendixcode,caption={Accepted revision that omits the required chloride component.}]{listings/cases/salt_C.json}
The self-proposed \op{OutcomeMatch} checks consistency with the cation the model chose; it cannot recover the omitted requirement. This is a specification gap. The example motivates retaining task obligations during revision and supporting the required representation.

\Needspace{12\baselineskip}
\subsection{Retrosynthesis: The Answer Is the Full Active Set}
\label{app:case-retro}
\begin{lstlisting}[style=appendixprompt,caption={Retrosynthesis task input.}]
Propose reactants that produce the following target via a acylation 
reaction.
Product: COc1ccc(C(N)=O)cc1C(=O)Nc1cccc(Cl)c1
Output reactant SMILES separated by dots.
\end{lstlisting}
The accepted one-pass trace cuts the amide connection between atoms 11 and 13 using \op{Fragment}. Both fragments remain active. It then attaches hydroxyl at the acyl-carbon anchor to obtain the carboxylic-acid precursor. The result consists of the acid and an aniline derivative, and the saved reactant set is correct.
\fixedlisting[style=appendixcode,caption={Complete accepted retrosynthetic trace.}]{listings/cases/retro_B.json}
The submitted \op{OutcomeMatch} assertion refers only to molecule 0, whereas reaction scoring compares the complete active set. Thus an accepted assertion about one fragment and a correct multi-molecule answer are distinct observations. \op{Fragment}, unlike \op{Eject}, retains the second fragment as part of that answer.

\Needspace{12\baselineskip}
\subsection{Functional-Group Generation: A Native Structural Obligation}
\label{app:case-amine}
\begin{lstlisting}[style=appendixprompt,caption={Functional-group generation task input.}]
The molecule is composed of 1 amine group.
\end{lstlisting}
The model proposes methylamine and asserts one amine group. Both the proposal and the native \op{FGCount} check are accepted, and the saved task outcome is successful. This compact example separates proposing a chemically valid candidate from stating the structural requirement that makes it relevant to the task. No refinement is needed.
\fixedlisting[style=appendixcode,caption={Complete accepted methylamine trace.}]{listings/cases/amine_B.json}

\FloatBarrier
\Needspace{12\baselineskip}

\section{Evaluation Details}
\label{app:tasks}
\subsection{Task Setup and Scoring}
Generation begins from an empty state; editing and optimization begin from the input molecule. Forward prediction begins from reactants and retrosynthesis from the product. Editing requires one active output molecule and disallows whole-molecule replacement, while optimization permits candidate proposals. Generation, ChEBI-20, and reaction tasks use terminal realizability checks. Closing bookkeeping applies throughout.

Non-reaction tasks score the first active molecule; reactions score the full active set. A rejected whole-trace completion can still provide a scored answer from its available state, or a parsed answer when no state is returned. Failed certification can therefore coexist with a correct answer. Missing or invalid answers receive zero. Selected episodes ending in context exhaustion are no-results, remain in the denominator, and receive zero score and certification; CoT and One-pass answers for the same example are scored independently.

ChEBI-20 uses its full supplied test split. USPTO selection uses $\operatorname{MD5}(\text{reaction key})\bmod100\geq90$, with a key combining patent identifier and row index. Forward and reverse versions of a reaction remain on the same side. We evaluate 500 examples per direction and supply reaction class names to all protocols. Optimization direction success is reported below separately from the similarity-weighted main-table score.

\begin{table}[htbp]
\centering
\caption{\textbf{Qwen3.6-27B optimization direction success (\%).} Each task contains 500 examples; input similarity is not included in this table.}
\label{tab:opt-success}
\small
\begin{tabular}{lrrrr}
\toprule
Objective & CoT & One-pass & Refine & MCP \\
\midrule
logP optimization & 59.8 & 95.8 & 99.8 & 99.4 \\
MR optimization & 60.6 & 93.2 & 99.8 & 99.6 \\
QED optimization & 54.2 & 65.6 & 92.2 & 77.0 \\
 
\bottomrule
\end{tabular}
\end{table}

\subsection{Inference Settings}
\label{app:settings}
Whole-trace refinement uses a 32,768-token context, temperature 0.3, up to 12,000 requested output tokens, and at most eight revisions after the initial one-pass completion. It stops at acceptance, its budget limit, or an early failure, with parse handling as described in Appendix~\ref{app:prompts}. The task reference is excluded from verifier feedback and used only for answer scoring.

MCP requests a 65,536-token context, temperature 0.7, and 4,096 output tokens per model call, with at most two attempts, 30 counted action/check calls, and 10 refusals per attempt. Read-only \op{measure} and \op{show\_state} calls and the closing \op{finish} call are outside that call counter. The first certified attempt is selected; otherwise the final attempt supplies the outcome.

\FloatBarrier
\Needspace{24\baselineskip}
\section{Additional Results}
\subsection{Whole-Trace Refinement Across Budgets}
\label{app:refinement-rounds}
Figure~\ref{fig:refinement-rounds} in Section~\ref{sec:experiments} shows the accepted-trace rate and answer score after at most $k$ whole-trace revisions, with $k=0$ denoting the shared One-pass completion. An accepted trace is retained at all larger budgets. The curves use the saved intermediate outcome at each budget, and the stopping rule uses only verifier feedback. Unsuccessful examples remain in each task's denominator. A context-exhausted continuation receives zero once its next completion is unavailable, while its independently saved earlier outcomes are retained at smaller budgets.

\subsection{Interaction-Length Distribution in MolLean-MCP}
\label{app:mcp-calls}
\begin{figure}[p]
\centering
\includegraphics[width=\linewidth]{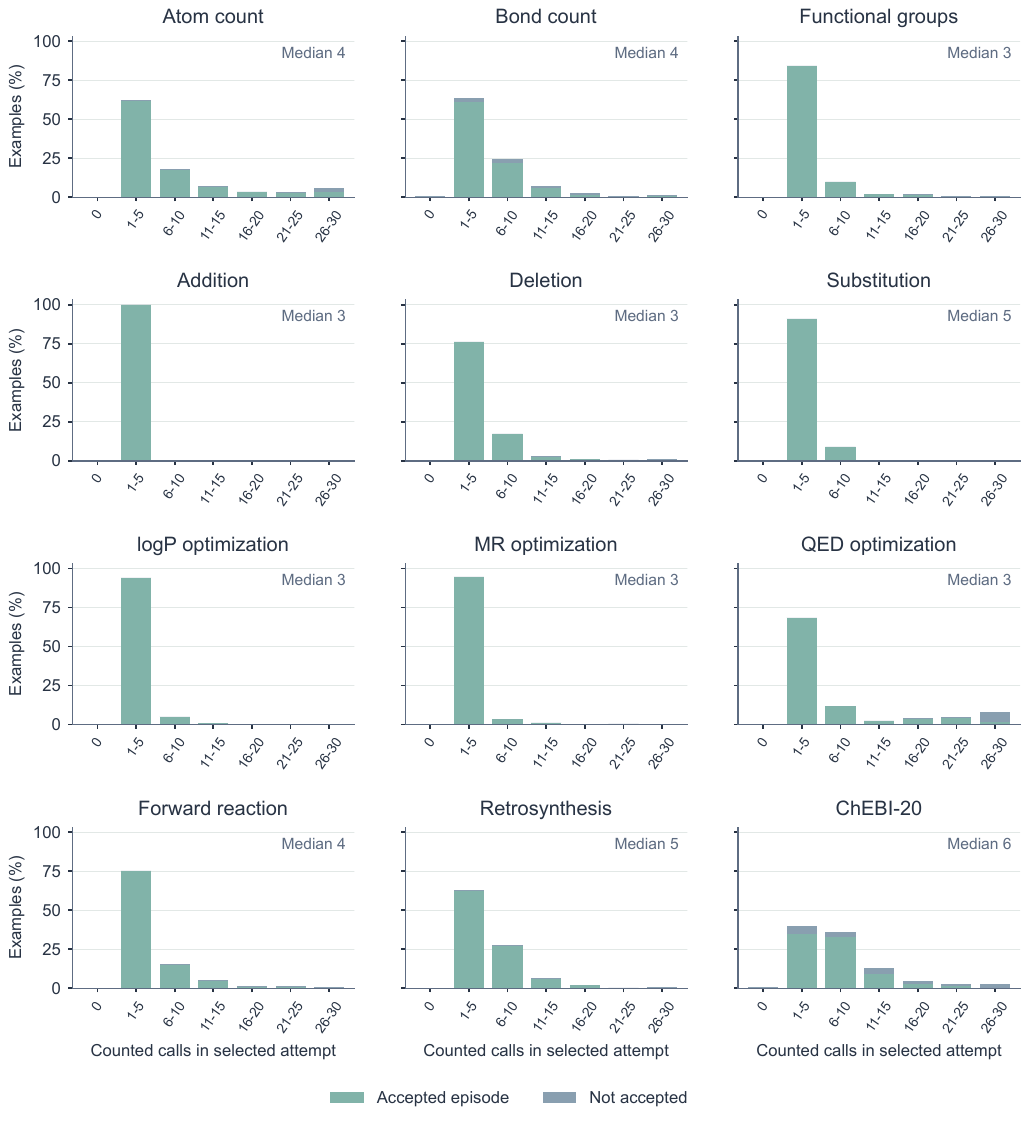}
\caption{\textbf{Distribution of counted MCP action/check calls in the selected attempt.} Bars sum to 100\% within each complete task cohort, including unaccepted and no-result examples. Colors separate accepted from unaccepted episodes. Each panel reports its median. Bins span zero, 1--5, 6--10, 11--15, 16--20, 21--25, and 26--30 calls. Counts exclude \op{measure}, \op{show\_state}, and \op{finish}, as well as earlier attempts.}
\label{fig:mcp-calls}
\end{figure}

Figure~\ref{fig:mcp-calls} describes the action/check budget consumed in the selected MCP attempt for every evaluated example. Selection follows the evaluation protocol: the first certified attempt, or the final attempt when none certifies. Counted calls include rejected actions and false assertions and exclude \op{measure}, \op{show\_state}, and \op{finish}.

The selected attempts are typically short on editing and descriptor optimization: the median is three calls for addition, deletion, logP, MR, and QED, and five for substitution. Reaction medians are four calls for forward prediction and five for retrosynthesis. ChEBI-20 has a longer interaction distribution, with a median of six and an interquartile range of four to ten calls. The tails also distinguish tasks: QED and atom-count generation contain appreciable mass beyond ten calls even though their medians are small. The accepted and unaccepted portions of each bar show whether the selected attempt closed successfully. Longer interaction can reflect additional construction, checking, or failed actions.

\FloatBarrier

\Needspace{22\baselineskip}
\subsection{Certification and Statistical Uncertainty}
\label{app:statistics}
Table~\ref{tab:verification} in Section~\ref{sec:experiments} reports trace acceptance and paired 95\% bootstrap intervals for the primary Refine-minus-CoT scores. Each interval uses 4,000 paired resamples within its task. The intervals describe variation across examples conditional on the model outputs.

\FloatBarrier

\Needspace{22\baselineskip}
\subsection{Joint and Paired Outcomes}
\label{app:joint}
Table~\ref{tab:joint} separates acceptance from binary task success for Refinement. Each row sums to the complete task denominator. For optimization, success here means improvement in the requested property direction.

\begin{table}[H]
\centering
\caption{\textbf{Joint outcomes for Qwen3.6-27B Refinement.}}
\label{tab:joint}
\small
\begin{tabular}{lrrrrr}
\toprule
& & \multicolumn{2}{c}{Sound} & \multicolumn{2}{c}{Not sound} \\
Task & $N$ & Success & Failure & Success & Failure \\
\midrule
Atom counts & 500 & 444 & 23 & 0 & 33 \\
Bond counts & 500 & 345 & 79 & 0 & 76 \\
Functional groups & 500 & 209 & 279 & 0 & 12 \\
Add component & 500 & 500 & 0 & 0 & 0 \\
Delete component & 500 & 485 & 14 & 0 & 1 \\
Substitute component & 500 & 480 & 19 & 0 & 1 \\
logP optimization & 500 & 499 & 1 & 0 & 0 \\
MR optimization & 500 & 499 & 0 & 0 & 1 \\
QED optimization & 500 & 458 & 12 & 3 & 27 \\
Forward reaction & 500 & 298 & 143 & 16 & 43 \\
Retrosynthesis & 500 & 96 & 345 & 8 & 51 \\
ChEBI-20 text to molecule & 3,300 & 1075 & 1533 & 17 & 675 \\
 
\bottomrule
\end{tabular}
\end{table}

\label{app:paired}
Table~\ref{tab:paired} pairs the same initial one-pass completion with its final refinement outcome. Recovered answers change from failure to success; lost answers change from success to failure. The other two columns retain their original status.

\begin{table}[H]
\centering
\caption{\textbf{One-pass to Refinement transitions for Qwen3.6-27B.} Each row sums to $N$; categories use binary task success.}
\label{tab:paired}
\small
\begin{tabular}{lrrrr}
\toprule
Task & Both succeed & Recovered & Lost & Both fail \\
\midrule
Atom counts & 307 & 137 & 0 & 56 \\
Bond counts & 199 & 146 & 0 & 155 \\
Functional groups & 194 & 15 & 0 & 291 \\
Add component & 468 & 32 & 0 & 0 \\
Delete component & 472 & 13 & 1 & 14 \\
Substitute component & 471 & 9 & 0 & 20 \\
logP optimization & 479 & 20 & 0 & 1 \\
MR optimization & 466 & 33 & 0 & 1 \\
QED optimization & 322 & 139 & 6 & 33 \\
Forward reaction & 107 & 207 & 11 & 175 \\
Retrosynthesis & 56 & 48 & 2 & 394 \\
ChEBI-20 text to molecule & 802 & 290 & 33 & 2175 \\
 
\bottomrule
\end{tabular}
\end{table}
\FloatBarrier

\Needspace{22\baselineskip}
\subsection{Qwen3-4B Whole-Trace Results}
\label{app:4b}
Qwen3-4B uses the same task IDs and whole-trace evaluation structure. Table~\ref{tab:4b} in Section~\ref{sec:experiments} reports CoT, One-pass, and Refinement; it does not include MCP experiments, as 4B models have a relatively poor interactive capability.

\FloatBarrier

\end{document}